# Coreference Resolution System for Indonesian Text with Mention Pair Method and Singleton Exclusion using Convolutional Neural Network

Turfa Auliarachman
School of Electrical Engineering and Informatics
Bandung Institute of Technology
Indonesia
t.auliarachman@gmail.com

Ayu Purwarianti
School of Electrical Engineering and Informatics
Bandung Institute of Technology
Indonesia
ayu@stei.itb.ac.id

*Abstract*—Neural network has shown promising performance on coreference resolution systems that uses mention pair method. With deep neural network, it can learn hidden and deep relations between two mentions. However, there is no work on coreference resolution for Indonesian text that uses this learning technique. The state-of-the-art system for Indonesian text only states the use of lexical and syntactic features can improve the existing coreference resolution system. In this paper, we propose a new coreference resolution system for Indonesian text with mention pair method that uses deep neural network to learn the relations of the two mentions. In addition to lexical and syntactic features, in order to learn the representation of the mentions' words and context, we use word embeddings and feed them to Convolutional Neural Network (CNN). Furthermore, we do singleton exclusion using singleton classifier component to prevent singleton mentions enter any entity clusters at the end. Achieving 67.37% without singleton exclusion, 63.27% with trained singleton classifier, and 75.95% with gold singleton classifier on CoNLL average F1 score, our proposed system outperforms the state-of-the-art system.

*Keywords—coreference resolution; deep learning; convolutional neural network; Indonesian language*

## I. Introduction

Coreference resolution is a task in Natural Language Processing (NLP) for determining linguistic expressions that refer to the same entity in the real world. In this paper, we call those linguistic expressions as mentions. For example, in the sentences "Alice sent a message to Bob. But he never received it.", "Bob"-'he' and "a message"-"it" are said to corefer. The purpose of coreference resolution system is to generate clusters of mentions which the members of those single clusters refer to the same entity in the real word.

The differences between Indonesian and English language characteristics introduce different problems in designing coreference resolution system. Unfortunately, the problems in Indonesian language are harder than in English language. For example, there is no definite noun and demonstrative noun in Indonesian language [1]. These two types of noun are intuitively easier to solve in this task. Other example is most of Indonesian words do not have gender information. Indonesian pronouns do not have gender information either. In English coreference resolution system, the gender of words and pronouns can be an easy way to increase or decrease the confidence of two mentions being corefer. Furthermore, there is no difference between subject pronouns, object pronouns, and possessive pronouns. For a machine learning model, it is easier to cluster "his chair" with "the chair" and exclude "he" than their Indonesian counterpart ("kursi dia" – "kursi" – "dia"). Last but not least, there is a special case with the suffix "-nya". The meaning of suffix "-nya" highly depends on the context of it. It can be used as a possessive pronoun, to make a noun become an adjective, to make a sentence sounds more polite, or any other several uses. Even as a possessive noun, "-nya" can be used for second person or third person.

There are at least two methods that can be used for supervised coreference resolution system, mention pair method and entity-mention method [2]. We choose the former because it is much easier to implement and much faster to train. Furthermore, to the best of our knowledge, all coreference resolution works for Indonesian text used the mention pair method.

In this paper, we first discuss some related works on coreference resolution system for Indonesian text [1], [3] and coreference resolution system that employs Convolutional Neural Network (CNN) . Then we describe our proposed coreference resolution system for Indonesian text. For training data, we use default training data generation method. We combine CNN-learned mention and its context representation with lexical and syntactic features and feed them to a deep neural network. Last, we do mention clustering with greedy best-first algorithm. We employ singleton classifier component in the system to prevent singleton mentions enter any entity clusters.

We evaluate our systems on the same dataset that is used by [1] using CoNLL average F1 score metrics [4]. Our systems outperform the state-of-the-art system [1] and one other coreference resolution system for Indonesian text [3]. Despite the low performance of our trained singleton classifier model, it can improve the performance of some of our bad performing systems.

## II. Related Works

There is only a few works on coreference resolution system for Indonesian text. To the best of our knowledge, [3] was the first coreference resolution work for Indonesian text. They used mention pair method in their work. Budi et al. [3] employed association rules learning algorithm, generating association rules with the left-hand side is the mention pair features and right-hand side is a boolean value denoting whether the mention pair corefer or not.

Suherik and Purwarianti [1] modified the former work with the addition of lexical and syntactic features. They also proposed a training data generation method which reduced the amount of training data and reduced the imbalance of those. They employed C45 decision tree algorithm and showed that the addition of lexical and syntactic features can improve the performance of the existing coreference resolution system. Because of this, we use all of their features on our work.

Moving on to related works on coreference resolution for English text, this work is most similar to [5]. They used deep neural network and CNN for learning mentions and their head words representations. We do not use head words representations as our feature because of the limited NLP tools for Indonesian language. They used hand-engineered features which we do not use because we would rather use [1]'s set of features which can be generated programmatically and gave good result specifically for Indonesian text. They also employed singleton classifier, but did not investigate whether this component was a good addition or not.

III. PROPOSED COREFERENCE RESOLUTION SYSTEM

A. Training Data Generation

Suherik and Purwarianti [1] showed that their training data generation method outperform [6]'s method and default method (which is pairing all mentions). However, we believe that because we use deep neural network, we need to feed more data to our coreference classifier models, and our coreference classifier models need to learn all the cases of mention pairs. Thus, we choose the default method, which pairs all mentions with other mentions. To make sure that our choice is not wrong, we also train coreference classifier models using [1]'s method and we compare both methods.

B. Features

We use the same exact features for singleton classifier and coreference classifier. Those features can be classified as three feature groups which are described below. However, we still conduct experiments on combination of the feature groups for singleton classifier as those features never have shown to be the best combination of feature groups for singleton classifier.

- **Mention's words**. We use word embeddings to represent the mention's words.

- **Context**. We use 10 words that preced and proceed the mention as context. Word embeddings are used to represent those context words.

- **Mention features**. For this feature group, we use 4 programmatically-generated features such as in [1]. Those features are (1) binary representation of whether the mention is a pronoun or not, (2) entity type, (3) binary representation of whether the mention is a proper name or not, and (4) binary representation of whether the mention is a first person pronoun or not.

- **Mention pair relation features**. For this feature group, we use 6 features such as in [1], with additional of 3 distance features correlated with coreferenceness. Those features are (1) binary representation of whether the two mentions are exact match or not, (2) binary representation of whether the two mentions have the same set of words or not, (3) binary representation of whether one mention is a substring of the other mention or not, (4) binary representation of whether one mention is an abbreviation of the other mention or not, (5) binary representation of whether the two mentions are appositives or not, (6) binary representation of whether the two mentions are the nearest candidate of each other or not, (7) sentence-based distance of the mentions, (8) word-based distance of the mentions, and (9) mention-based distance of the mentions. This feature group is only used in coreference classifier.

For word embeddings, we use 300 dimensional pre-trained word embeddings. The word embeddings were trained using word2vec [7] algorithm.

C. Neural Network Architecture

We create two deep neural network models for this experiment, that are singleton classifier and coreference classifier. Fig. 1 is the architecture of our singleton classifier and Fig. 2 is the architecture of our coreference resolution.

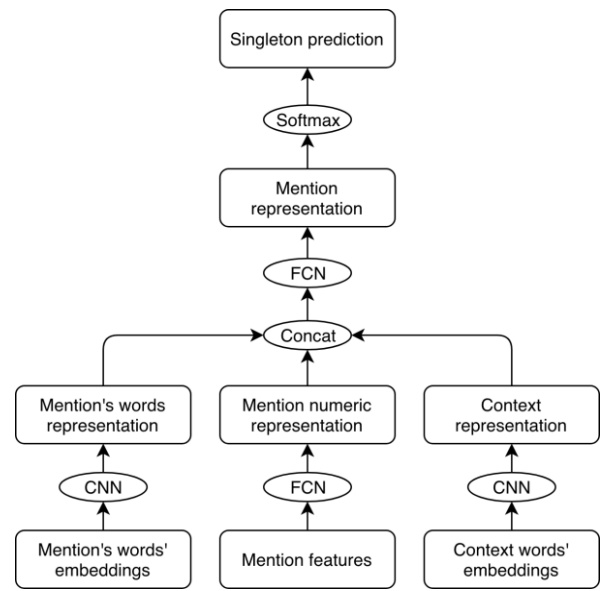

Fig. 1 Singleton classifier architecture.

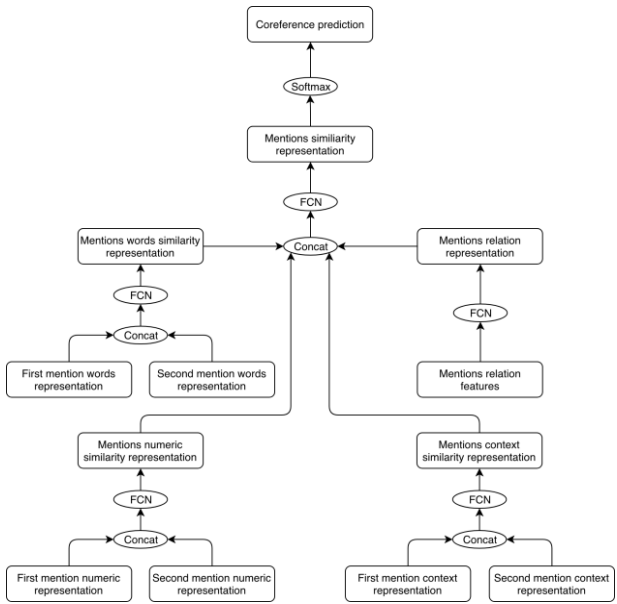

Fig. 2 Coreference classifier architecture.

At its core, the architecture of those two models are similar. We build representations of each feature groups and concatenate them to make a higher level representation. In coreference classifier, each feature group representations of each mentions are concatenated first to make similarity representation of the feature group. After every concatenations, we employ Fully Connected Network (FCN) to learn the representation of those concatenations.

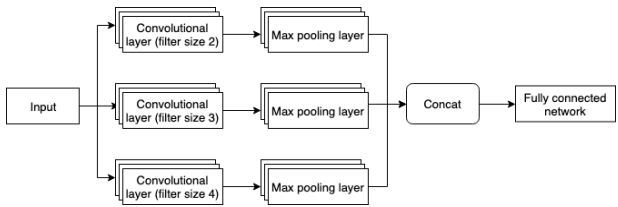

Fig. 3 Convolutional neural network.

All the CNN networks in our proposed architectures consist of exactly the same architecture as shown in Fig. 3. They have 64 filters each for every size of 3 filter sizes (2, 3, 4), which are pooled with max pooling. However, we also try [5]'s hyperparameters as explained in the next section.

The difference between our architectures and the ones in [5] is that they used mentions dependency head words. We do not use this feature due to the limited NLP tools for Indonesian language. Instead, we use context words to capture what are being discussed when the mentions appeared. One other difference is they added an additional representation layer for embeddings-related feature groups. We do not add this additional layer because we believe that it is enough to directly concatenate all the feature groups representations.

### D. Neural Network Hyperparameters

Wu and Ma [5] proposed a set of hyperparameters configuration to use in CNN and FCN. They proposed 5 layers of each CNN, using 200 filters of size 2. They proposed 5 layers of each FCN except for the 10 layers of final FCN. They used 200 dimensions for each of their hidden layers.

As for our Indonesian coreference resolution, we analyzed that the FCN layer numbers and dimensions mentioned in [5] are too excessive for our case since we only employ a small number of features. Thus, we propose the use of 3 different filter sizes for each CNN, creating 64 filters for each sizes. For FCN that are directly connected to the inputs, we propose 2 layers with dimensions 32 and 16. For every FCN after concatenations, we propose 3 layers with dimensions 64, 32, and 16. Finally, for the final FCN, we propose 2 layers with dimensions 32 and 8. In our experiment, we compare both hyperparameters configurations.

### E. Mentions Clustering

Mentions clustering is the last process in our proposed coreference resolution system. Because the output of our coreference classifier is the confidence of a mention pair being coreference, we use greedy best-first algorithm for clustering our mentions. In this algorithm, each mentions are placed in the same disjoint set with a mention that it produce the maximum confidence score. Those disjoint sets are the output of our system. We also do singleton exclusion in this phase. That is, if a mention is predicted to be a singleton, it will be skipped when the algorithm searches for the maximum confidence score for any mention.

## IV. EXPERIMENTS

Our systems were evaluated on the same dataset with [1]. Originally, it consisted of 1,030 news sentences. However, they only used 759 sentences for training and 108 sentences for testing. We extracted 811 non-singleton mentions and 4,221 singleton mentions for training data, plus 105 non-singleton mentions and 798 singleton mentions for testing data. We evaluated our system using CoNLL average F1 score metrics [4], which computes the average F1 score from three popular metrics: MUC [8], $B^3$ [9], and CEAFe [10]. We compared the performance of our systems with [1] and [3].

### A. Singleton Classifier Experiment

First we conducted an experiment on singleton classifier to get the best model. The best model was then used in the next experiment on coreference resolution system. We evaluated our singleton classifier models using weighted F1 metrics.

In this experiment, we evaluated the best combination of hyperparameter configuration and feature groups for singleton classifier. There were 2 hyperparameter configurations to consider: configuration of [5] and our proposed configuration. Combined with 7 combination of feature groups, we had 14 models to evaluate. Below are the 7 combinations of feature groups.

1) Mention's words,
2) Context,
3) Mention features and mention pair relation features,
4) Mention's words and context,
5) Mention's words, mention features, and mention pair relation features,

6) Context, mention features, and mention pair relation features,
7) All feature groups.

TABLE I. SINGLETON CLASSIFIER EXPERIMENT RESULT

| Hyperparameters Configuration | Feature Groups | Non-Singleton F1 | Singleton F1 | Weighted F1 |
|---|---|---|---|---|
| Wu and Ma [5] | 1 | 68% | 94% | 91% |
| Wu and Ma [5] | 2 | 49% | 94% | 88% |
| Wu and Ma [5] | 3 | 69% | 94% | 91% |
| Wu and Ma [5] | 4 | 67% | 95% | 91% |
| Wu and Ma [5] | 5 | 68% | 95% | 91% |
| Wu and Ma [5] | 6 | 70% | 95% | 92% |
| Wu and Ma [5] | 7 | **71%** | **96%** | **93%** |
| Proposed | 1 | 68% | 95% | 91% |
| Proposed | 2 | 46% | 93% | 87% |
| Proposed | 3 | 69% | 94% | 91% |
| Proposed | 4 | 67% | 95% | 92% |
| Proposed | 5 | 69% | 95% | 92% |
| Proposed | 6 | 68% | 94% | 91% |
| Proposed | 7 | 62% | 95% | 91% |

From Table I above, we can see that the best performance was achieved when we used [5]'s hyperparameters configuration using all the features available. However, its performance on non-singleton class was still low, only gaining 71% F1 score.

Our best model performed poorly on predicting "*-nya*" mentions. It only gained 75.51% accuracy for "*-nya*" mentions. This model also performed poorly on predicting proper names, only gained 76.44% accuracy for proper name mentions.

B. *Coreference Resolution System Experiment*

For sanity check, we first created a random classifier and observed its performance. This is to make sure that our systems really solve the coreference resolution problem. Table II shows that even the best random classifier performed poorly for this problem. It is because each predictions correlate one with another.

TABLE II. PERFORMANCE OF RANDOM CLASSIFIER

| | Precision | 9.96 |
|---|---|---|
| MUC | Recall | 100.00 |
| | F1 | 18.12 |
| | Precision | 1.12 |
| $B^3$ | Recall | 100.00 |
| | F1 | 2.21 |
| | Precision | 1.61 |
| CEAFe | Recall | 0.33 |
| | F1 | 0.66 |
| CoNLL Average F1 | | 7.00 |

We conducted the coreference resolution system experiment in two variables for coreference classifier: training data generation method and neural network hyperparameters configuration. There were also three variables for the use of singleton classifier in this experiment: without singleton classifier, with trained singleton classifier, and with gold singleton classifier. Gold singleton classifier was used as we wanted to see how far we could improve our system if we had a good singleton classifier.

Table III below shows the experiment result with [5]'s hyperparameters configuration, whereas Table III shows the experiment result with our proposed hyperparameters configuration.

TABLE III. COREFERENCE RESOLUTION SYSTEM EXPERIMENT RESULT WITH WU AND MA [5] HYPERPARAMETERS CONFIGURATION

| Training Data Generation | Singleton Classifier | MUC | | | $B^3$ | | | CEAFe | | | CoNLL Average F1 |
|---|---|---|---|---|---|---|---|---|---|---|---|
| | | Precision | Recall | F1 | Precision | Recall | F1 | Precision | Recall | F1 | |
| Budi et al. [3] | | 41.46 | 65.38 | 50.75 | 33.74 | 55.55 | 41.98 | 32.48 | 60.14 | 42.18 | 44.97 |
| Suherik and Purwarianti [1] | | 63.95 | 70.51 | 67.07 | 50.41 | 63.80 | 56.32 | 57.26 | 57.26 | 57.26 | 60.22 |
| Default | No | 28.83 | 82.05 | 42.67 | 10.33 | 79.72 | 18.29 | 33.53 | 23.60 | 27.70 | 29.55 |
| Default | Trained | 61.33 | 58.97 | 60.13 | 42.57 | 53.02 | 47.23 | 72.60 | 37.64 | 49.58 | 52.31 |
| Default | Gold | 87.01 | 85.90 | 86.45 | 62.13 | 83.22 | 71.14 | 84.82 | 43.98 | 57.92 | 71.84 |
| Suherik and Purwarianti [1] | No | 39.05 | 84.62 | 53.44 | 17.56 | 81.08 | 28.87 | 50.10 | 31.54 | 38.71 | 40.34 |
| Suherik and Purwarianti [1] | Trained | 63.01 | 58.97 | 60.93 | 45.83 | 51.69 | 48.58 | 66.23 | 39.25 | 49.29 | 52.93 |
| Suherik and Purwarianti [1] | Gold | 84.93 | 79.49 | 82.12 | 64.36 | 75.23 | 69.37 | 82.47 | 48.87 | 61.37 | 70.95 |

TABLE IV. COREFERENCE RESOLUTION SYSTEM EXPERIMENT RESULT WITH PROPOSED HYPERPARAMETERS CONFIGURATION

| Training Data Generation | Singleton Classifier | MUC | | | $B^3$ | | | CEAFe | | | CoNLL Average F1 |
|---|---|---|---|---|---|---|---|---|---|---|---|
| | | Precision | Recall | F1 | Precision | Recall | F1 | Precision | Recall | F1 | |
| Budi et al. [3] | | 41.46 | 65.38 | 50.75 | 33.74 | 55.55 | 41.98 | 32.48 | 60.14 | 42.18 | 44.97 |
| Suherik and Purwarianti [1] | | 63.95 | 70.51 | 67.07 | 50.41 | 63.80 | 56.32 | 57.26 | 57.26 | 57.26 | 60.22 |
| Default | No | 73.33 | 70.51 | 71.90 | 60.33 | 65.26 | 62.70 | 77.17 | 60.02 | 67.52 | 67.37 |
| Default | Trained | 84.62 | 56.41 | 67.69 | 74.55 | 50.15 | 59.96 | 77.69 | 51.79 | 62.15 | 63.27 |
| Default | Gold | 94.83 | 70.51 | 80.88 | 86.46 | 64.79 | 74.07 | 88.25 | 62.10 | 72.90 | 75.95 |
| Suherik and Purwarianti [1] | No | 54.47 | 85.90 | 66.67 | 35.73 | 81.81 | 49.74 | 52.32 | 50.39 | 51.34 | 55.91 |

| Training Data Generation | Singleton Classifier | MUC | | | B³ | | | CEAFe | | | CoNLL Average F1 |
|---|---|---|---|---|---|---|---|---|---|---|---|
| | | Precision | Recall | F1 | Precision | Recall | F1 | Precision | Recall | F1 | |
| Budi et al. [3] | | 41.46 | 65.38 | 50.75 | 33.74 | 55.55 | 41.98 | 32.48 | 60.14 | 42.18 | 44.97 |
| Suherik and Purwarianti [1] | | 63.95 | 70.51 | 67.07 | 50.41 | 63.80 | 56.32 | 57.26 | 57.26 | 57.26 | 60.22 |
| Suherik and Purwarianti [1] | Trained | 68.25 | 55.13 | 60.99 | 56.22 | 48.44 | 52.04 | 62.60 | 46.37 | 53.27 | 55.43 |
| Suherik and Purwarianti [1] | Gold | 88.31 | 87.18 | 87.74 | 70.84 | 83.88 | 76.81 | 87.88 | 55.33 | 67.91 | 77.49 |

Table III and Table IV shows that our proposed hyperparameters configuration performed significantly better than [5]'s hyperparameters configuration on our neural network. The default training data generation performed better than [1]'s on our proposed configuration, whereas [1]'s performed better than the default method on [5]'s configuration. That being said, we can not conclude which training data generation method is better from our experiment.

On the poor-performing systems, the use of our trained singleton classifier could improve their performance. This improvement could be as high as 22.76% (29.55% to 52.31%, see Table III). However, in the good-performing system, the use of our trained singleton classifier rather decreased its performance by 4.10% (67.37% - 63.27%, see Table IV). We suspect that it was caused by the low performance of our trained singleton classifier on non-singleton class.

The interesting part is when the gold singleton classifier was used, the system with average F1 score 55.91% outperformed the system with average F1 score 67.37%. Furthermore, the system with average F1 score 29.55% gained competitive performance with other much better systems. This can be the prove that singleton exclusion can unleash the hidden potential of low performing systems.

## V. CONCLUSION

Our experiments showed that our proposed system with CNN and singleton exclusion process outperformed the state-of-the-art system [1]. Our best model achieved 67.37% (without singleton exclusion), 63.27% (with trained singleton classifier), and 75.95% (with gold singleton classifier) on CoNLL average F1 metrics, whereas the state-of-the-art system [1] achieved 60.22%.

Singleton exclusion process can significantly improve the performance of coreference resolution systems, with a condition that the singleton classifier has a good performance. Even our low performing trained singleton classifier improved a system which initially gained 29.55% by 22.76%. That being said, in the future, we aim to do works on building singleton classifier models using other topologies and set of features to get better singleton classifier models.

REFERENCES

[1] G. J. Suherik and A. Purwarianti, "Experiments on coreference resolution for Indonesian language with lexical and shallow syntactic features," in 2017 5th International Conference on Information and Communication Technology (ICoICT), 2017.

[2] J. Zheng, W. W. Chapman, R. S. Crowley, and G. K. Savova, "Coreference resolution: A review of general methodologies and applications in the clinical domain," J. Biomed. Inform., pp. 1113–1122, 2011.

[3] I. Budi, S. Bressan, and Nasrullah, "Co-Reference Resolution for the Indonesian Language Using Association Rules.," iiWAS, pp. 117–125, 2006.

[4] S. Pradhan, A. Moschitti, N. Xue, O. Uryupina, and Y. Zhang, "{CoNLL-2012} Shared Task: Modeling Multilingual Unrestricted Coreference in {OntoNotes}," in Proceedings of the Sixteenth Conference on Computational Natural Language Learning (CoNLL 2012), 2012.

[5] J. Wu and W. Ma, "A Deep Learning Framework for Coreference Resolution Based on Convolutional Neural Network," in 2017 IEEE 11th International Conference on Semantic Computing (ICSC), 2017, pp. 61–64.

[6] W. M. Soon, D. C. Y. Lim, and H. T. Ng, "A machine learning approach to coreference resolution of noun phrases," Comput. Linguist., 2001.

[7] T. Mikolov, K. Chen, G. . Corrado, and J. Dean, "Efficient Estimation of Word Representations in Vector Space," Proc. Work. ICLR, vol. 2013, 2013.

[8] M. Vilain, J. Burger, J. Aberdeen, D. Connolly, and L. Hirschman, "A Model-Theoretic Coreference Scoring Scheme," in Message Understanding Conference (MUC), 1995.

[9] A. Bagga and B. Baldwin, "Algorithms for Scoring Coreference Chains," in The first international conference on language resources and evaluation workshop on linguistics coreference, 1998, pp. 563–566.

[10] X. Luo, "On coreference resolution performance metrics," in Proceedings of the Conference on Human Language Technology and Empirical Methods in Natural Language Processing, 2005, pp. 25–32.